\documentclass[letterpaper, 10 pt, conference]{ieeeconf}  % Comment this line out if you need a4paper

\IEEEoverridecommandlockouts                              % This command is only needed if 
\usepackage{graphics} % for pdf, bitmapped graphics files
\usepackage{epsfig} % for postscript graphics files
\usepackage{times} % assumes new font selection scheme installed
\usepackage{amsmath} % assumes amsmath package installed
\usepackage{amssymb}  % assumes amsmath package installed
\usepackage{booktabs}
\usepackage{float}
\usepackage{algorithm}
\usepackage[noend]{algpseudocode}

\newcommand{\oursnsp}{{\text STRUG}}
\newcommand{\ours}{{\text STRUG }}

\title{\LARGE \bf
Task-Directed Exploration in Continuous POMDPs \\for  Robotic Manipulation of Articulated Objects
}

\author{Aidan Curtis$^{1}$, Leslie Kaelbling$^{1}$, Siddarth Jain$^{2}$% <-this % stops a space
\thanks{$^{1}$Computer Science and Artificial Intelligence Laboratory, Massachusetts Institute of Technology, Cambridge, MA, USA. This research was completed during A. Curtis’s internship at MERL.}
\thanks{$^{2}$Mitsubishi Electric Research Laboratories (MERL), Cambridge, MA, USA. 
        {\tt\small sjain@merl.com }}%
}

\begin{document}

\maketitle
\thispagestyle{empty}
\pagestyle{empty}

%%%%%%%%%%%%%%%%%%%%%%%%%%%%%%%%%%%%%%%%%%%%%%%%%%%%%%%%%%%%%%%%%%%%%%%%%%%%%%%%
\begin{abstract}

Representing and reasoning about uncertainty is crucial for autonomous agents acting in partially observable environments with noisy sensors. Partially observable Markov decision processes (POMDPs) serve as a general framework for representing problems in which uncertainty is an important factor. Online sample-based POMDP methods have emerged as efficient approaches to solving large POMDPs and have been shown to extend to continuous domains. However, these solutions struggle to find long-horizon plans in problems with significant uncertainty. Exploration heuristics can help guide planning, but many real-world settings contain significant task-irrelevant uncertainty that might distract from the task objective. In this paper, we propose \oursnsp, an online POMDP solver capable of handling domains that require long-horizon planning with significant task-relevant and task-irrelevant uncertainty. We demonstrate our solution on several temporally extended versions of toy POMDP problems as well as robotic manipulation of articulated objects using a neural perception frontend to construct a distribution of possible models. Our results show that \ours outperforms the current sample-based online POMDP solvers on several tasks. 

\end{abstract}

%%%%%%%%%%%%%%%%%%%%%%%%%%%%%%%%%%%%%%%%%%%%%%%%%%%%%%%%%%%%%%%%%%%%%%%%%%%%%%%%
\section{Introduction}

\begin{figure}[h]
\centering
    \includegraphics[width=1.0\linewidth]{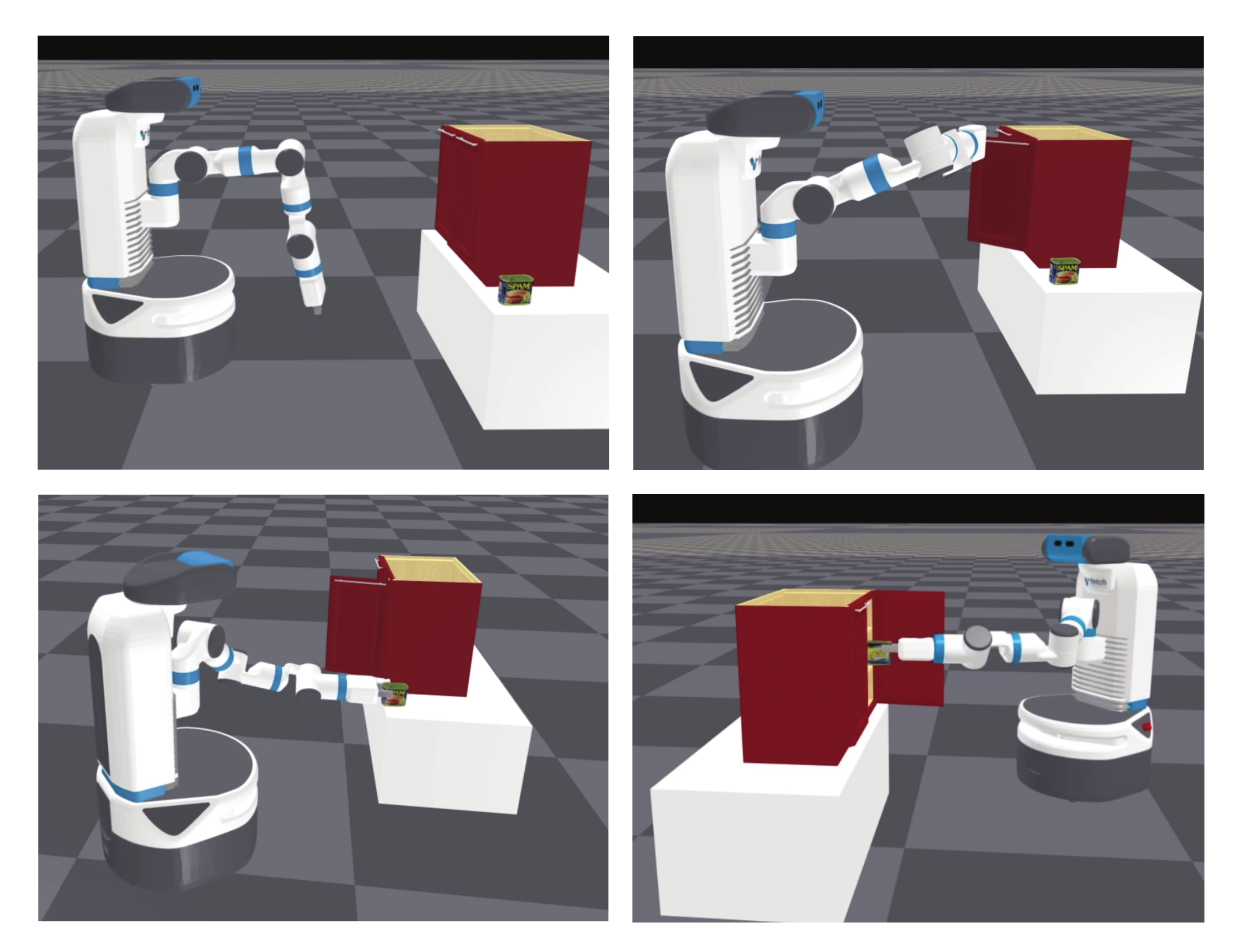}
    \caption{Visualization of an articulated object robotic manipulation task (\textsc{OpenDoor}) with snapshots of the robot looking at the articulated object (top left), opening the cabinet door (top right), picking up the target object (bottom left), and placing the object inside the cabinet (bottom right).}
    \label{fig:robots}
\vspace{-0.5cm}
\end{figure}

Typical model-based approaches to robotics make decisions based on the most likely state of the world. While a point estimate of the world state is tolerable for some applications, it is not sufficient for problems that require actions to improve the model estimate, or in cases where optimal plans under incorrect model assumptions have adverse and irreversible effects. 
For instance, the robot in Figure~\ref{fig:robots} has a head-mounted camera that can noisily estimate the state of a cabinet object. The robot also has access to controllers (e.g., \texttt{OpenDrawer}) that it can use to affect the world. Executing one of those controllers under an incorrect world model could lead to unintended effects, such as pulling the cabinet off the table or breaking something. An optimal strategy might be to partially execute a controller, observe the effect, and then execute the correct controller under the new estimated model. Finding such an optimal plan requires a representation of model uncertainty and a decision-making strategy that reasons about uncertainty.

Decision-making under uncertainty is a widely studied topic that spans many research areas~\cite{probroboticsbook} including motion planning~\cite{uav_motion}, control theory \cite{control_uncertainty}, reinforcement learning~\cite{brl}, and task and motion planning~\cite{BHPN, SSReplan}. 
The most general formulation of the problem is as a partially observable Markov decision process (POMDP). 
Unfortunately, finding optimal solutions to POMDPs can be computationally intractable. Online POMDP solvers have an advantage over optimal solvers because they consider only sections of the belief space reachable from the current state, which enables them to be applied to larger state spaces. 
Online POMDP solvers such as POMCP~\cite{POMCP} and DESPOT~\cite{DESPOT} that use a particle-based representation of uncertainty have proven to be the most efficient in large state spaces and have even been applied to problems with continuous observations and actions~\cite{POMCPOW}.   

Although these approaches are capable of working in large and continuous state spaces, they struggle with long-horizon planning when information-gathering is necessary for reducing model uncertainty due to the doubly exponential nature of the search tree. These issues can sometimes be overcome by augmenting the reward or regularizing the value function to explicitly incentivize uncertainty reduction~\cite{ENTREG_POMDP}. 
Other approaches propose active learning, exclusively for uncertainty reduction, followed by maximum likelihood model-based planning~\cite{POMDPLite}, 
which works when all of the uncertainty about the model is relevant to the task. 
However, for many real-world tasks, there is a significant amount of uncertainty about components of the state that are irrelevant to the task. 
A robot that tries to reduce all uncertainty in the task shown in Figure~\ref{fig:robots} will interact with all of the drawers to learn each joint parameter, even when only one drawer is necessary.
% gain information Given a cabinet, the robot may perform open and close actions on all drawers to gather information. 
% Accomplishing such tasks involves repeated sense-plan-act phases under uncertainty in the robot’s observations of the state. 
% One focus of our work hereby lies in manipulating articulated objects with task-relevant uncertainty guidance. For example, if the agent in Figure 1 is given the task ``Place an object in the top drawer'', it only needs to reduce uncertainty about one of the drawers. 

In this paper, we propose Search with Task Relevant Uncertainty Guidance (\oursnsp), a particle-based online POMDP solver capable of finding long-horizon paths to the goal with significant task-relevant and task-irrelevant uncertainty. \ours uses a domain-independent strategy for identifying task-relevant information-gathering actions. At a high level, \ours relies on a fast heuristic-based forward search planner to find per-particle plans and then uses these plans to identify subgoals that reduce model uncertainty.

Our work presents two key contributions. First, we develop a novel metric for task-relevant uncertainty and show usefulness in the context of an online belief tree policy search. Second, we demonstrate the effectiveness of our approach by showing how robotic manipulation tasks with complex kinematics can be modeled as a POMDP with information gathering actions and solved using task-relevant uncertainty.

%%%%%%%%%%%%%%%%%%%%%%%%%%%%%%%%%%%%%%%%%%%%%%%%%%%%%%%%%%%%%%%%%%%%%%%%%%%%%%%%
\section{Related Work}
\label{sec:related_work}

The topics of planning under uncertainty and estimating articulated models for robot manipulation are widely studied. 

\textbf{Online POMDP solvers:} Online POMDP solvers have been shown to be effective in high-dimensional state spaces in comparison to other POMDP solvers~\cite{POMCP, DESPOT, 7139503}.
Recent extensions have shown that these approaches also work in continuous action and observation spaces \cite{POMCPOW} with several applications to robotics~\cite{robot_manip_pomdp, Chen2016POMDPliteFR}.

\textbf{Planning with uncertainty:} 
One approach to planning with uncertainty is to model the uncertainty on various state variables using parameterized continuous distributions and define how certain actions will affect various statistics on those distributions~\cite{bhpnn}. 
While this approach works well with structured uncertainty and known action effects, it fails in the presence of unstructured uncertainty. 
Another way of dealing with uncertainty is to first reduce any model ambiguity through active learning and then plan in the resulting model \cite{al2, al1}.
Other approaches try to merge the model learning and planning steps using entropy regularized objective functions or reward augmentation to encourage information-gathering~\cite{ENTREG_POMDP, POMDPLite}. Although active learning and entropy-regularized approaches to planning with uncertainty can effectively reduce general uncertainty, they are overwhelmed when there is significant goal-irrelevant uncertainty.

\textbf{Kinematic Model Learning:}
Many approaches from computer vision estimate articulated object models from a sequence of observations \cite{RPMNet, pairs}.
While these methods produce impressive results, it is unclear where these image sequences would come from in autonomous robotic systems. Other approaches have attempted to estimate kinematic models from single images \cite{opd, pmlr-v100-abbatematteo20a, AOPE, sagci}. 
These approaches are more applicable but are error-prone due to partial visibility and the ambiguity inherent in still images of dynamic objects. 
Some works have experimented with model-free learning for interaction with articulated objects \cite{learning1, Mo_2021_ICCV}, but these approaches tend to be less efficient in real-world interactions and can be potentially dangerous in safety-critical situations.
The most closely related works to ours model kinematic uncertainty and take actions to reduce that uncertainty \cite{4543220, al2, honda}. To our knowledge, we are the first to attempt to reduce \textit{task-relevant} uncertainty for estimating kinematic models in the context of a task.
%%%%%%%%%%%%%%%%%%%%%%%%%%%%%%%%%%%%%%%%%%%%%%%%%%%%%%%%%%%%%%%%%%%%%%%%%%%%%%%%
\section{Background}
\label{sec:background}

In this section, we outline the POMDP formulation, describe one class of algorithms used in large and continuous state spaces, and show an extension to the classic formulation that allows for belief-dependent search guidance.

\subsection{POMDPs}
A standard infinite-horizon POMDP is defined by the tuple $\langle{\mathcal{S}, \mathcal{O}, \mathcal{A}, \mathcal{T}, \mathcal{R}, \mathcal{Z}, \gamma}\rangle$ where $\mathcal{S}, \mathcal{O}, \mathcal{A}$ are the state, observation, and actions spaces respectively, $\mathcal{T}$ defines the transition probabilities that govern state transitions $P(s_{t+1}| s, a)$, $\mathcal{R}$ defines the reward function $\mathcal{R}(s, a)$, $\mathcal{Z}$ defines the observation function $P(o_{t+1}|s_t, a_t)$, and $\gamma$ is the discount factor. 

While this is the standard formulation used for offline solvers, online Monte-Carlo simulation methods use a generative model POMDP definition $\langle{G, \mathcal{R}, b_0, \gamma}\rangle$ where $G$ is a generative function $G(s_t, a_t)$ that yields $(s_{t+1}, o_{t+1})$ drawn from $P(s_{t+1}, o_{t+1}|s_{t}, a_{t})$, and $b_0$ defines the initial state distribution $P(s_0)$. 
The initial belief $b_0$ exists in a belief space $\mathcal{B}$, which is the space of probability distributions over the state space. 
This generative formulation lifts the restrictive assumption that the $\mathcal{T}$ can be represented in closed form. 
A solution to a POMDP is a policy $\pi:\mathcal{B}\rightarrow{\mathcal{A}}$ that maps beliefs in $\mathcal{B}$ to actions in $\mathcal{A}$ such that the value of the initial belief is maximized, with value defined as follows 

\begin{equation}
V_\pi(b) = {\mathbb{E}}\Big[\sum_{t=0}^{\infty}\gamma^t R(s_t, \pi(b_t))\Big]\;.
\end{equation}

\subsection{Belief Tree Policy Search}

One class of approaches attempts to maximize $V_{\pi}(b_0)$ by building a belief tree with $b_0$ as the root node. 
Nodes in this belief tree alternate between belief nodes and observation nodes. 
Multiple actions can be taken from a belief node and multiple observations branch from a single observation node due to the stochasticity of the model and uncertainty in the underlying state.
At leaf nodes of the belief tree, $V_{\pi}(b)$ is estimated by simulating action sequences using a random or heuristic-guided policy. 
At intermediate nodes of the belief tree, $V_{\pi}(b)$ is estimated from its children by taking an argmax over possible actions and an expectation over resulting observations. 
The value of an optimal policy $\pi^*$ at belief $b$ is then defined as follows

\begin{equation}
V_{\pi^*}(b) = \underset{a\in{\mathcal{A}}}{\text{max}}\big( \mathbb{E}_{b}[R(s, a)] + \gamma\sum_{o\in{\mathcal{O}}} P(o|b, a) V_{\pi^*}(b') \big) \;.
\end{equation}

% \sidd{both side have $V^*$? also should just say that ``the value of belief state $b$ is set equal to the immediate reward for taking the best action for $b$ plus the discounted expected value of the resulting belief state $b'$"}
% \aidan{Yes, it's defined recursively as is typical (Check out DESPOT paper). Are you saying this description should replace the equation?}

Where $b'$ is the result of performing a belief update using action $a$, observation $o$, and prior belief $b$. 
In settings with particle belief representations in which a belief is sets of possible states, the belief update is simply $b' = \{ s_{t+1}\; | \; s_{t+1} , o_{t} = G(s_t, a_t), o = o_{t}\}$. 
Sample-based POMDP solvers typically build this belief tree incrementally in an approach similar to MCTS \cite{POMCP, POMCPOW, DESPOT}. 
Given a belief tree, a policy tree can be extracted by selecting the action that maximizes value at each belief node, leaving a tree that only branches on observations. 

\subsection{Belief-Dependent Rewards}
An important variant of this problem, and the one that we use in this paper, formulates the reward to be a function of the belief $\mathcal{R}(s, a, b)$ where b is computed from a history of action, observation pairs \cite{araya2010}. 
This formulation has been used to augment the reward to encourage information-gathering actions that reduce entropy in the agent's belief \cite{POMDPLite, Dressel_Kochenderfer_2017}. 

Active learning approaches directly optimize for reduction in belief state entropy $\mathbb{E}[log(b_s)]$. 
Other task-directed POMDP solvers augment the reward function with the belief state entropy to encourage actions that lead to low uncertainty along a trajectory \cite{ENTREG_POMDP, entropy, POMDPLite}:

\begin{equation}
\label{eq:belief_entropy_reduction}
\mathcal{\hat{R}}(s, a, b) = \mathcal{R}(s, a)+\beta\mathbb{E}[log(b)]\;, 
\end{equation}

% \begin{equation}
% \label{eq:belief_entropy_reduction}
% \text{Maximize}\;\;\;{\color[rgb]{0.620814,0.044661,0.040054}\mathcal{R}(s, a)}\;
% \end{equation}

% \begin{equation}
% \label{eq:belief_entropy_reduction}
% \text{Maximize}\;\;\;{\color[rgb]{0.620814,0.044661,0.040054}\mathcal{R}(s, a)}+{\beta\color[rgb]{0.117241,0.508649,0.763023}\mathbb{E}[log(b)]}\;
% \end{equation}
% $\sim$
% \begin{equation}
% \label{eq:belief_entropy_reduction}
% \text{Maximize}\;\;\;\beta{\color[rgb]{0.117241,0.508649,0.763023}\mathbb{E}[log(b)]} \;\;\text{then}\;\;\;{\color[rgb]{0.620814,0.044661,0.040054}\mathcal{R}(s, a)}
% \end{equation}

% \begin{equation}
% \label{eq:belief_entropy_reduction}
% \text{Maximize}\;\;\;{\color[rgb]{0.620814,0.044661,0.040054}\mathcal{R}(s, a)}+\beta{{\color[rgb]{0.574939,0.107606,0.725653}\text{TRU}(b)}}\;
% \end{equation}

% \begin{equation}
% \label{eq:belief_entropy_reduction}
% \mathcal{\hat{R}}(s, a, b) = \mathcal{R}(s, a)+\beta\mathbb{E}[log(b)]\;, 
% \end{equation}
% \aidan{get rid, say its a baseline}
where $\beta$ is a hyperparameter that determines the exploration exploitation tradeoff. Our approach formulates a new augmented belief-dependent reward function that includes a measure of task-relevant uncertainty.

%%%%%%%%%%%%%%%%%%%%%%%%%%%%%%%%%%%%%%%%%%%%%%%%%%%%%%%%%%%%%%%%%%%%%%%%%%%%%%%%

\section{Method}
\label{sec:method}
This section describes our method for solving POMDPs with belief-dependent reward functions.
We call our approach \ours for \textit{search with task-relevant uncertainty guidance}.
At a high level, \ours samples particles, or possible states, from the initial belief and uses an uncertainty-free heuristic planner to find maximum-reward plans for each particle. 
The particle-specific plans are then evaluated on the other particles sampled from the initial state distribution yielding a probability of success for each particle's plan on each other particle. 
We then run a search in belief space with an augmented reward function that incentivizes actions leading to observations that separate particles with mutually incompatible plans. 
Our key insight is that dissimilar states in the state space often share successful plans, especially when there is a significant degree of task-irrelevant uncertainty. 

\subsection{Task-Relevant Uncertainty}
We first define task-relevant uncertainty (TRU) of a policy to be the expected variance in value under trajectories sampled from the policy and transition model.
We want to incentivize our search to find a policy $\pi$ with low task-relevant uncertainty (TRU) defined

% \[\text{TRU} = \mathbb{E}_{\mathcal{T}}\Big[\Big(V_\pi(b^\prime)-\mathbb{E}_{\mathcal{T}}\big[V_\pi(b^{\prime\prime})\big]\Big)^2\Big]\]
\begin{equation}
\text{TRU}(b) = \mathbb{E}_{s_1\sim{b}}\Big[\text{Var}_{s_2\sim{b}}[V_{\pi^*_{s_1}}(s_2)]\Big]\;.
\end{equation}
This is notably different from energy-based or distance-based objectives that express uncertainty in state. 
A reasonable approach for estimating this objective during planning is to sample a number of candidate action sequences with a random policy, evaluate the value for each simulated action sequence, and find the sample variance. 

However, random actions are unlikely to obtain any reward in sparse-reward environments, resulting in a TRU estimate of zero even when TRU is high. 
A more accurate TRU estimate in sparse environments requires a denser sampling in the support of the reward function. 
To obtain this denser sampling, we can identify a set of action sequences with high likelihood of reaching the goal for some particular particle using an uncertainty-free planner. 
Subsequent sections describe how this metric is calculated and used in the context of an online sample-based POMDP solver for both discrete and continuous action and observation spaces.

\begin{figure*}[h]
\centering
    \includegraphics[width=1.0\linewidth]{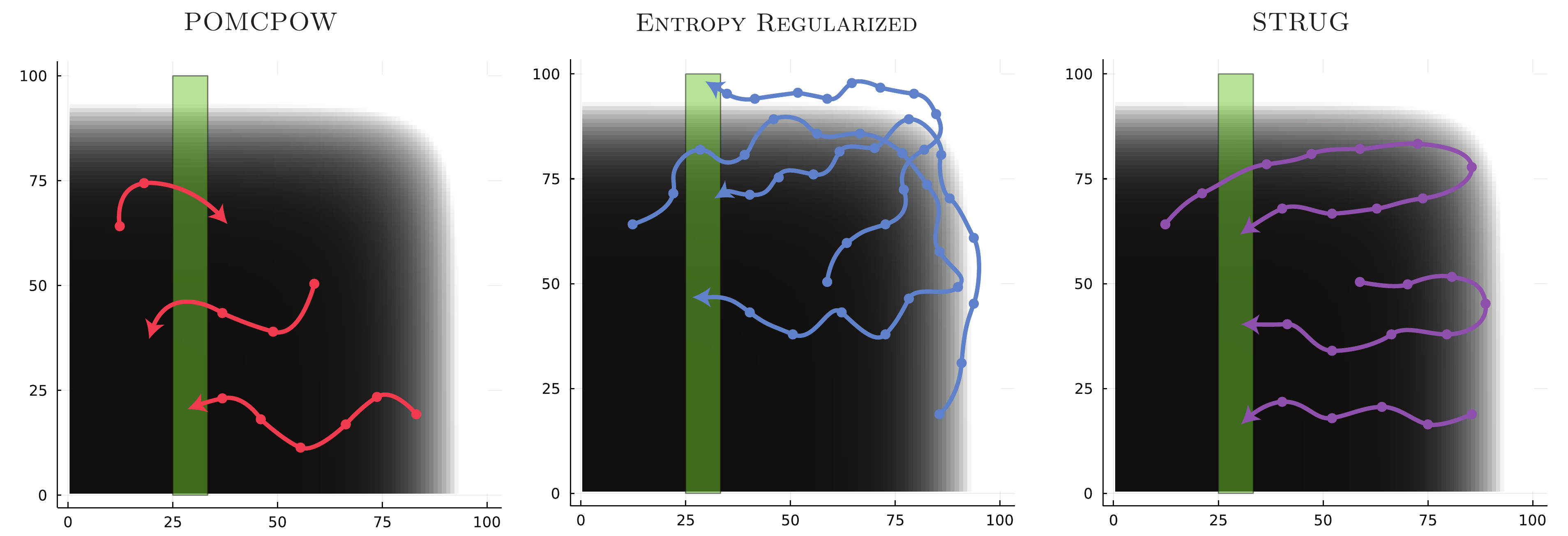}
    \caption{Visualization of sample trajectories from three POMDP solvers on the LightDark2D Task. The agent starts with uncertainty in its position, and has to navigate to the goal region (vertical green bar). The top wall reduces uncertainty about the y position, and the right wall reduces uncertainty about the x position. \textsc{POMCPOW} fails to take any information-gathering actions, \textsc{Entropy Regularized} reduces all state uncertainty before going to the goal. \ours reduces only task-relevant uncertainty before reaching the goal.}
    \label{fig:ld2d}
    \vspace{-0.3cm}
\end{figure*}

\subsection{\ours Algorithm}
We start by sampling a set of $M$ particles $X = \{x^0_0, ..., x^p_0\}$ from the initial belief state $b_0$.
A determinized search is then performed to obtain a maximum-reward plan $\bar{a}^i = ((a^i_0, x^i_0), ..., (a^i_T, x^i_T))$ on each particle $x^i_0\in{X}$. 
We then estimate the expected value of each plan on each sampled particle $x_j\in{X}$ by executing plan $\bar{a}^i$ $K$ times from state $x_j$, inducing $K$ state trajectory rollouts $(x^{i, j}_0, ..., x^{i,j}_{T})_k$. 
This  results in an $M\times{M}$ $\textit{plan compatibility matrix}$
\begin{equation}
S_b^{i,j} = \frac{1}{K}\sum_{k=0}^{K}{\sum_{t=0}^{T}\gamma^t\mathcal{R}(x_t^{i, j})}\;.
\end{equation}

See Algorithm~\ref{alg:compatibility_matrix} for details. Element $(i, j)$ in matrix $S_b$ indicates the expected value of executing particle i's optimal plan in particle j. 
It follows that TRU is calculated from $S_b$ in the following way:
\begin{equation}
\text{TRU}_{S_b}(b) = \sum_{j=0}^M b(x_0^j)\sum_{i=0}^M \Big(S_b^{i, j}-\sum_{j^\prime=0}^M b(x_0^{j'})S_b^{i, j^\prime}\Big)^2\;.
\end{equation}

In order to apply our TRU estimate to an online tree-based POMDP solver, we need a way of calculating the \textit{incremental} TRU, or the TRU gain from taking a certain action in a certain belief state.
Incremental TRU is defined recursively as a function of the current belief state and action:

\begin{equation}
\Delta\text{TRU}(b, a) = \mathbb{E}_{o\in{\mathcal {O}}}\Big[\text{TRU}_{S_{b'}}(b')-\text{TRU}_{S_b}(b)\Big]\;.
\end{equation}

Unfortunately, calculating TRU at every belief node in the belief tree is computationally burdensome, with the primary bottleneck being per-particle planning time and quadratic plan evaluation time. 
To resolve this, we make an approximation of $\Delta\text{TRU}$ that uses a cached version of the plan compatibility matrix $S_{b_0}$ instead of recomputing it at each belief node during search. Approximate $\Delta\text{TRU}$ is defined as
\begin{equation}
\Delta\widehat{\text{TRU}}(b, a) = \mathbb{E}_{o\in{\mathcal {O}}}\Big[\text{TRU}_{S_{b_0}}(b')-\text{TRU}_{S_{b_0}}(b)\Big]\;\;.
\end{equation}
% \begin{equation}
% \label{eq:re}
% \Delta\widehat{\text{TRU}}(b, a) = \mathbb{E}_{o\in{\mathcal {O}}}\Big[\widehat{\text{TRU}}_{S_{b'}}(b')-\widehat{\text{TRU}}_{S_{b}}(b)\Big]
% \end{equation}
% $S_{\cdot, \mathcal{O}_j}$ with $\mathcal{O}_j = \{j \;|\; G(b_{t}[j], a_t) = (b_{{t+1}}[j], o) \}$, $n_o=|\mathcal{O}_j|$. 
Intuitively, we want to take actions that result in information that would be useful at the initial belief state. This approximation makes two important assumptions. First is that the uncertainty in the problem can be represented by a possibly infinite set of latent parameters that do not change throughout the problem \cite{POMDPLite, hipmdp}. The second is that the actions that gather information pertaining to the unchanging hidden variables do not lead to irreversible effects from which the goal cannot be achieved. For example, actions such as dropping a vase on the floor to see if it is made of glass would violate this assumption.
% This approximation makes the assumption that for a given action-observation pair, there exists no two states with non-zero probability in the belief that go from having incompatible optimal plans to compatible optimal plans or vice-versa with identical observations. 
% For example, an environment with irreversible and invisible action effects may have this property. 
% We have found $\Delta\widehat{\text{TRU}}$ to be a good practical approximation of $\Delta\text{TRU}$ in our experiments on a wide array of problems. 
The original reward function is augmented with the exploration reward to yield the final belief-dependent reward:
\begin{equation}
\label{eq:re}
\hat{\mathcal{R}}(s_t, a_t, b_t) = \mathcal{R}(s_t, a_t)+\beta\;\Delta\widehat{\text{TRU}}(b_t, a_t) \;,
\end{equation}

\noindent where $\beta$ is a hyperparameter specifying the importance of gathering task-relevant information. The full algorithm description including TRU augmentation can be seen in ~\ref{alg:policy}
In our experiments, we err on the side of selecting large importance values ($\beta=10$) because the TRU term will approach zero after all task-relevant information is gathered.

\begin{algorithm}[h]
  \caption{\textsc{CompatibilityMatrix}}
  \label{alg:compatibility_matrix}
  \begin{algorithmic}[1] % The number tells where the line numbering should start
    \Require \text{POMDP model p}
    \State $P \gets$ Sample $M$ particles $b \sim b_0$
    \State $F\gets\{\}, D \gets \{\}, S\gets{\text{zeros}(M, M)}$
    \For{$m \in 1:M$}
        \State $D \gets D \cup \textsc{Plan}(P[m])$
    \EndFor
    \For{$m \in 1:M$}
        \For{$d_i\in 1:|D|$}
            \For{$k\in 1:K$}
                \State $R \gets 0$
                \For{$a\in \text{reverse}(D[d_i])$}
                    \State $s^\prime, o, r \gets G(s, a)$
                    \State $R\gets{\gamma R+r}$
                \EndFor
                \State $S[m,d_i]\gets{S[m,d_i]+R/K}$
            \EndFor
        \EndFor
    \EndFor
    \Return $S$
   \end{algorithmic}
\end{algorithm}

\begin{algorithm}[h]
  \caption{\ours}
  \label{alg:policy}
  \begin{algorithmic}[1] % The number tells where the line numbering should start
    \Require \text{POMDP model p, Initial belief distribution $b_0$}
    
    \Procedure{Solve}{$p, b_0$}
        \State $S \gets \textsc{CompatibilityMatrix}(p)$
        \State $t \gets \textsc{EmptyTree}$
        \For{$i \in \text{tree\_samples}$}
            \State $\textsc{Simulate}(t, p, b_0)$        
        \EndFor
        \State \Return $\underset{a}{\mathrm{argmax}}\, t.Q(b_0, a)$
    \EndProcedure
    
    \Procedure{Sim}{$t, p, b$}
        \State $s \gets \text{sample}(b)$
        \State $a \gets \textsc{HPW}(s, p.A, t.N)$
        \State $b^\prime, o, r \gets G(b, a)$
        \State $s^\prime \gets \text{sample}(b^\prime)$
        \If{$t.N(b) = 1$}
            \State $r\gets r+\gamma\textsc{Rollout}(s^\prime)+\beta\Delta\widehat{TRU}(b^\prime, a)$
        \Else
            \State $r\gets r+\gamma\textsc{Sim}(t, p, b^\prime)+\beta\Delta\widehat{TRU}(b^\prime, a)$
        \EndIf
        \State $t.Q(s, a) \gets t.Q(s, a) + (r-t.Q(s^\prime, a))/t.N(s^\prime, a)$
        \State \Return $r$
    \EndProcedure
   \end{algorithmic}
\end{algorithm}

\begin{algorithm}[h]
  \caption{\textsc{HPW}}
  \label{alg:hpw}
  \begin{algorithmic}[1] % The number tells where the line numbering should start
    \Require \text{State $s$, action schemas $A$, state-action counts $N$}
    \State $a\gets{[]}$
    \For{$\Theta \in A.\text{discrete}$}
        \State $a \gets a \oplus \underset{a_i \in \Theta}{\mathrm{argmax}}\;\,{\text{UCB}(s, a_i)}$
    \EndFor
    \State $a_d\gets{a}$
    \For{$\Theta \in A.\text{continuous}$}
        \If{$|\Theta_s[a_d]|\leq kN(s, a)^\alpha$}
            \State $\Theta_s[a_d] \gets \Theta_s[a_d] \cup \text{sample}(\Theta)$
        \EndIf
        \State $a \gets a \oplus \underset{a_i\in \Theta_s[a_d]}{\mathrm{argmax}}\,{\text{UCB}(s, a_i)}$
    \EndFor
    \Return $a$
   \end{algorithmic}
\end{algorithm}

\subsection{Hierarchical Progressive Widening}
Sampling continuous actions during tree search is typically performed using Progressive widening (PW), double progressive widening (DPW), or Voronoi progressive widening (VPW) \cite{upperconfidencetrees, POMCPOW, vpw}. These all work by limiting the number of child nodes of a certain parent node. For example, progressive widening limits the number of child action nodes to $kN^\alpha$ where $N$ is the number of samples reaching the parent node, and $k, \alpha$ are hyperparameters of the search process.
This strategy does not easily extend from fixed-dimensional purely continuous action spaces to the hybrid discrete-continuous actions spaces that are typical in robotic planning applications. Typical robotic planning applications define a number of controllers $a\in{\mathcal{A}}$, each with a unique set of continuous parameters $\theta_a$ \cite{tamp_survey}. 
A naive extension of double progressive widening might use a fixed-size continuous search space with dimensionality $\max_{a\in{\mathcal{A}}}|\theta_a |$, ignoring the extraneous continuous parameters for controllers with fewer parameters. Unfortunately, this approach is highly inefficient due to redundancy. We propose a new Hierarchical Progressive Widening (HPW) strategy that splits action selection into a sequence of decisions about discrete action parameters followed by a joint selection of the continuous parameters with progressive widening constraints. 
HPW uses the standard UCB criterion at each step of the action selection process, giving priority to exploration of the continuous parameters of promising controllers without unnecessary redundancy \cite{mcts_survey}:

\begin{equation}
\text{UCB}(s, a) = Q(s, a) + c\;\sqrt{\frac{\text{log}\; N(s)}{N(s, a)}}\;.
\end{equation}
$Q(s, a)$ is the estimated value of an action in a particular state, N(s) is the state sample count, $N(s, a)$ is the state-action sample counts, and $c$ is the exploration parameter. In our experiments we set $c$ to be the maximum reward. See Algorithm~\ref{alg:hpw} for details.

\begin{figure*}[h]
\centering
    \includegraphics[width=1.0\linewidth]{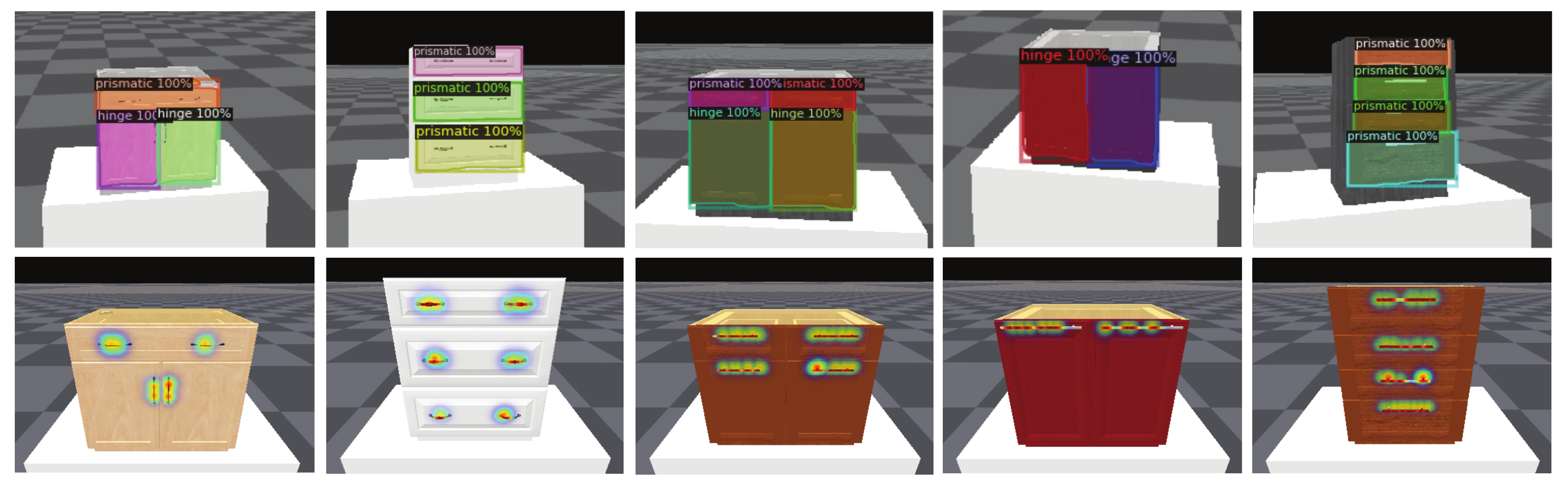}
    \caption{A visualization of the perceptual model results used to construct the POMDP problem. Top: Results of the MaskRCNN detection module for link mask prediction. Bottom: The combined, thresholded, and filtered Where2act heatmap for the Pull controller}
    \label{fig:perception}
\end{figure*}

\begin{table*}[h]
\centering
\caption{Experimental results on the tasks described in Section~\ref{sec:experiments}. The mean discounted reward and standard error across 100 seeds. Rewards are between -10 and 10. Scores within 2 points of the highest-scoring algorithm are bolded}
\label{table:results}
\begin{tabular}{@{}ccccccccc@{}}
\toprule
             & Tiger               & ExtendedTiger       & LightDark 1D       & LightDark 2D       & Shelf               & Open Drawer             & Open Door  \\ \midrule
Random       & $-1.62\pm0.98$      & $-0.21\pm0.89$      & $-5.43\pm0.65$     & $-3.13\pm0.14$     & $-9.34\pm0.20$      & $-9.47\pm0.05$     & $-9.54\pm0.04$     \\
POMCP (Disc) & $\bf{5.67\pm0.75}$  & $-0.02\pm0.98$      & $1.30\pm0.98$      & $-0.57\pm0.12$     & $\bf{9.02\pm0.00}$  & $-1.00\pm0.33$     & $-1.50\pm0.41$     \\
POMCPOW      & $\bf{6.49\pm0.66}$  & $-0.37\pm0.81$      & $2.60\pm0.97$      & $-1.20\pm0.89$     & $\bf{8.62\pm0.19}$  & $1.30\pm0.71$      & $0.20\pm0.71$      \\
Entropy Regularized   & $\bf{6.25\pm0.31}$  & $\bf{3.95\pm0.34}$  & $0.12\pm0.94$      & $1.26\pm0.41$      & $3.91\pm0.11$       & $0.34\pm1.18$      & $-0.15\pm0.97$      \\
Active Learning  & $\bf{5.59\pm0.28}$  & $\bf{3.96\pm0.33}$  & $\bf{5.50\pm0.82}$ & $\bf{2.13\pm1.29}$      & $1.62\pm0.92$       & $1.49\pm1.91$      & $1.03\pm0.84$      \\
\ours        & $\bf{6.34\pm0.31}$  & $\bf{2.31\pm0.25}$  & $\bf{5.80\pm0.62}$ & $\bf{3.55\pm0.91}$ & $\bf{9.02\pm0.00}$  & $\bf{4.22\pm1.60}$ & $\bf{5.21\pm1.39}$ \\ \bottomrule
\end{tabular}
\end{table*}

\section{Experiments}
\label{sec:experiments}

Our baselines include planning methods that handle uncertainty and aim to trade off information-gathering with reward exploitation. Namely, we evaluate \textbf{POMCP}~\cite{POMCP} and its continuous variant \textbf{POMCPOW}~\cite{POMCPOW}. 
To apply POMCP to continuous problem domains, we followed the procedure described in \cite{POMCPOW} to discretize the action and observation spaces. 
The \textbf{Entropy Regularized} uses the augmented reward in Equation~\ref{eq:re} to explicitly encourage entropy reduction. 
Lastly, \textbf{Active Learning} separates the problem into two steps: an active learning step that explicitly aims to minimize model entropy and an exploitation step that uses MCTS to find an optimal plan under the most likely model. 

We perform experiments on several temporally extended versions of toy POMDP problems and on tasks involving robotic manipulation of articulated objects that relies on perception with a neural
network for object-detection to construct a distribution of possible model states. We evaluate on the following tasks that all require some form of information-gathering in addition to goal-directed actions to achieve good performance. For tasks involving robotic manipulation, we use the Fetch robot~\cite{Wise2016FetchF} in the Issac gym simulation~\cite{isaacgym}. 

\begin{itemize}
  \item \textbf{Tiger \cite{tiger}: }  The agent stands in front of two doors. Opening one door reveals a tiger (negative reward), and the other reveals treasure (positive reward). The waiting action incurs a slight negative reward but results in a noisy observation of the tiger's location. 
  \item \textbf{ExtendedTiger: } A modified version of the Tiger problem that requires several repeated wait actions before any observations are received.
  \item \textbf{LightDark1D/2D \cite{lightdark}: } The agent is uncertain about its current position, and has to navigate to a goal region.  It can gain information about its position by moving toward the light region(s). In the 2D environment, moving to particular walls reduce uncertainty about specific dimensions of the agent's location (see Fig. \ref{fig:ld2d}).
  \item \textbf{Shelf Manipulation:} The robot is placed in front of a piece of storage furniture (shelf) from the Partnet Mobility dataset \cite{pn_mobility} with both hinge and door joints and a spam object is placed on a table from the YCB object set. A reward is received when the object is placed inside the storage furniture. Although the agent is uncertain about many aspects of the shelf model, it does not need to reduce uncertainty to achieve the goal.
  \item \textbf{Open Drawer (Door): } The same setup as the Shelf problem, but the articulated object has changed to only have prismatic (hinge) joints.  The agent is uncertain about many aspects but only needs to reduce uncertainty about a single drawer (door) to achieve the goal.
%   \item \textbf{Object Search: } Find an object in the storage Furniture. The belief is constructed by sampling possible object positions in the model of the storage furniture estimated from perception. 
\end{itemize}

\subsection{Robotics Task Details}
\subsubsection{Dynamics \& Reward}
Our simulated experiments were performed in the Pybullet~\cite{coumans2016pybullet} physics simulator and visualized in IsaacGym~\cite{isaacgym}. The physics simulator served as the dynamics function $f$ of our POMDP. 
As shown in Figure~\ref{fig:robots}, we use a mobile-base Fetch robot model~\cite{Wise2016FetchF} with articulated object models from the Sapien Partnet Mobility dataset~\cite{pn_mobility}. Initial state distributions are generated by sampling multiple camera perspectives with the robot RGBD camera and passing the images through a MaskRCNN object detection model that predicts articulated object link masks, joint types, and joint parameters~\cite{opd}. If the base of the kinematic model moved more than $0.1$ meters, the environment terminated, and a reward of -10 was returned. If the task was completed successfully the environment was terminated, and a reward of 10 was given. A maximum of 10 steps was allowed.
\subsubsection{Initial State Distribution}
The MaskRCNN was fine-tuned from the Detectron2 model~\cite{wu2019detectron2} using simulated data captured in IsaacGym on objects from the Partnet Mobility dataset not used during evaluation. 
Because the trained model had a tendency to be confidently incorrect, we additionally added noise to the input images, class predictions, and output joint parameters. 
Example bounding boxes, and predicted masks can be seen in Figure~\ref{fig:perception}\;(top). 
\subsubsection{Observation Function}
The observation function is defined using the Hausdorff distance, $\text{H}$ between the expected and received pointcloud observation:
\begin{equation}
\mathcal{Z}(o|s, a) = \mathcal{N}\big(\text{H}(o, \textsc{pcd}(f(s, a)), \sigma^2\big)\;,
\end{equation}
where $\textsc{pcd}$ is the simulated observation that results from execution of action a on state s. We used $\sigma=10^{-3}$.

\subsubsection{Action Space}
The actions consisted of parameterized controllers $\texttt{OpenDrawer(?g)}$, $\texttt{OpenDoor(?g)}$, $\texttt{Pick(?o)}$,  $\texttt{Place(?o, ?p)}$, $\texttt{Push(?g)}$, $\texttt{PullV(?g)}$, $\texttt{PullH(?g)}$ where $\texttt{?o}$ is the discrete object to manipulate, $\texttt{?p}$ is a 6 DoF object pose, and $\texttt{?g}$ is a 6 DoF gripper pose. 
To increase the potential for the pull controllers to impact the kinematic model state, we biased the controller parameter sampling using the Where2Act model~\cite{Mo_2021_ICCV}. 
The Where2Act models generate heatmaps of pull locations for horizontal and vertical orientations that highlight handles, buttons, and other likely object interfaces on unseen articulated objects. 
The Where2Act models are also trained from a simulated dataset using Partnet Mobility models. 
An example combined horizontal and vertical pull heatmap is shown in Fig.~\ref{fig:perception}\;(bottom). 

\subsection{Results \& Discussion}
Our results show all POMDP methods work on the simplest Tiger problem with relatively equal performance. In the more difficult ExtendedTiger problem, we observe that only approaches that perform explicit uncertainty reduction (Entropy Regularized, Active Learning, and STRUG) perform well. A similar trend is apparent in the LightDark 1D and LightDark 2D domains. We see the Entropy Regularized baseline struggle in the 1D version of this domain, and both Entropy Reduction and Active Learning perform worse in the 2D version. A few qualitative results shown in Figure~\ref{fig:ld2d} help elucidate the problem. Methods that reduce general uncertainty perform better than standard POMDP solvers but are less efficient than STRUG, which reduces only task-relevant uncertainty. For the shelf robotics task, we can see that the standard POMDP methods solve it effortlessly since no additional information is needed to solve the goal. However, the methods that reduce general uncertainty are less efficient because they try to reduce uncertainty anyway. Lastly, in the Open Drawer and Open Door robotics tasks, we observe that only STRUG is capable of solving them. Standard POMDP methods fail to consider information-gathering actions due to the temporally extended nature of the problem, and general information-gathering approaches spend more time gathering unnecessary information. 

Although our task selection does not highlight them, \ours does come with a set of assumptions and general limitations. First, it cannot be used in environments with irreversible actions that gain task-relevant information. Second, using an approximation of $\text{TRU}$ restricts \ours to environments in which the uncertain variables do not change during the task. In future work, we hope to address some of these shortcomings and move to larger task and motion planning robotics domains with more controllers, objects, and temporal dependencies.

\section{Conclusions}
\label{sec:conclusion}
In this paper, we present \oursnsp, an approach for planning with model uncertainty that prioritizes information-gathering actions the reduce task-relevant uncertainty. 
While directly reducing task-relevant uncertainty is intractable, we found a practical approximation under some assumptions. 
We evaluated this approach on a  number of POMDP problems, including robotic manipulation of articulated objects, a task that contained model uncertainty arising naturally from limitations in the robot's perception. 
Our findings show that augmenting reward with task-relevant uncertainty improves performance over existing POMDP solvers in temporally extended domains and outperforms approaches that reduce general uncertainty instead of task-relevant uncertainty.

\bibliographystyle{IEEEtran}
\typeout{} % hack to overcome some overleaf/latex bug
\bibliography{references}

\end{document}